\documentclass{article}

\usepackage{microtype}
\usepackage{graphicx}
\usepackage{subfigure}
\usepackage{booktabs} 

\usepackage{hyperref}

\usepackage[accepted]{icml2021}

\begin{document}

\twocolumn[
\icmltitle{How to Accelerate Capsule Convolutions in Capsule Networks}



\icmlsetsymbol{equal}{*}

\begin{icmlauthorlist}
\icmlauthor{Zhenhua Chen}{iu}
\icmlauthor{Xiwen Li}{wustl} 
\icmlauthor{Qian Lou}{iu} 
\icmlauthor{David Crandall}{iu}
\end{icmlauthorlist}

\icmlaffiliation{iu}{School of Informatics, Computing, and Engineering Indiana University, Bloomington, IN, USA}
\icmlaffiliation{wustl}{Washington University in St. Louis, USA}

\icmlcorrespondingauthor{Zhenhua Chen}{chen478@iu.edu}

]



\printAffiliationsAndNotice{}  

\begin{abstract}
The routing procedure in CapsNets is one of the bottlenecks that impact its efficiency. How to break this bottleneck has been studied a lot. However, the other bottleneck, the capsule convolution itself, has largely been neglected. Capsule convolution, as a special type of convolution, use capsules (tensors) rather than neurons (scalars) as the basic computation unit. All current deep learning frameworks are optimized based on general convolution. As a result, these optimizations do not work well for capsule convolutions. In this paper, we found that capsule convolutions can be considered as `multiplication of multiple small matrics' compared to `one big matrix multiplication' in convolution. Based on this observation, we develop two acceleration schemes with CUDA APIs and test them on a custom CapsNet. The result shows that our solution achieves a 4X times acceleration.  
\end{abstract}

\section{Introduction}

\subsection{Accelerating CapsNets}
CapsNets~\cite{dyrouting, emrouting} have drawn a lot of attention since proposed. A general structure of CapsNets is a hybrid of convolutional layers and capsule layers with a routing procedure. One of the bottlenecks in CapsNets is the routing procedure, and most of the current research is about either accelerating classic routing procedures~\cite{fast_dyrouting} or proposing new routing procedures~\cite{attention_routing, encapsule}. Besides routing procedures, the other vital structure of CapsNets is the capsule layers (capsule convolution layer, fully-connected capsule layer). Here we only discuss capsule convolution layer since fully-connected capsule layer is a special type of capsule convolution layer. Currently, most deep learning frameworks are designed to optimize general convolutions, thus they usually under-perform on capsule convolutions. In this paper, we first compare the difference between capsule convolution and general convolution, then based on our observations, we propose two possible solutions. In section~\ref{section:acce}, we give details on how our acceleration algorithm works. Finally, we show the acceleration performance on a custom CapsNet. 

\begin {figure*}[t]
\begin{center} 
\includegraphics[width=0.6\textwidth]{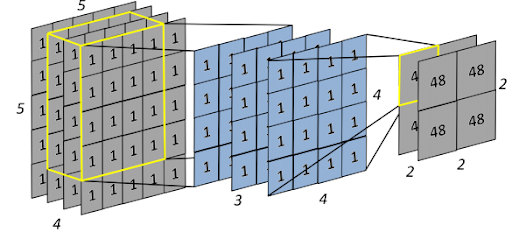}
\caption{3D convolution: tensor-to-scalar mapping. The shape of input is $5\times5\times4$. The shape of 3D kernel is $4\times4\times3$. As a result, the shape of output is $2\times2\times2$. Yellow area shows current input area being convolved by kernel and corresponding output.}
\label{3d_conv}
\includegraphics[width=.7\textwidth]{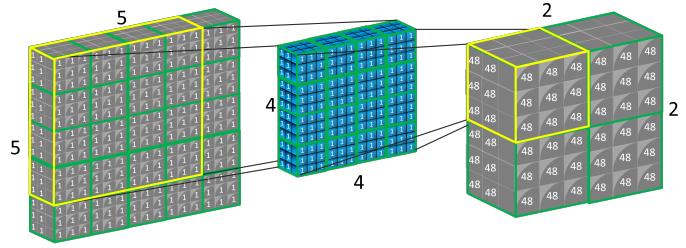}
\caption{Capsule convolution in P-CapsNets: tensor-to-tensor
  mapping. The input is a tensor of 1's which has a shape of $1\times
  5\times5\times(3\times3\times3)$ (correspond to the the input
  channel, input height, input width, first capsule dimension, 
  second capsule dimension, and third capsule dimension,
  respectively). The capsule kernel is also a tensor of 1's which has
  a shape of $4\times4\times1\times1\times(3\times3\times3)$ ---
  kernel height, kernel width, number of input channel, number of
  output channel, and the three dimensions of the 3D capsule. As a
  result, we get an output tensor of 48's which has a shape of
  $1\times2\times2\times(3\times3\times3)$. Yellow areas show current
  input area being convolved by kernel and corresponding output.}
\label{pcapsnet_structure} 
\end{center}
\end{figure*}

\subsection{Convolution versus capsule convolution }
Convolution is a mathematical operation between kernel and input. Roughly speaking, assume that we have an input $I(W, H, C)$ with width $W$, height $H$ and channel $C$ at each location $(x, y)$. Also, we have a kernel $K(w, h, c)$ with width $w$, height $h$ and channel $c$. A general convolution can be calculated as Algorithm~\ref{alg:conv}: 

\begin{algorithm}[tb] 
   \caption{Convolution.}
   \label{alg:conv}
\begin{algorithmic}
   \STATE {\bfseries Input, Kernel:} $I(W, H, C)$, $K(w, h, c)$
   \FOR{$i=1$ {\bfseries to} $1..W$}
      \FOR{$j=1$ {\bfseries to} $1..H$}
         \FOR{$p=1$ {\bfseries to} $1..w$}
            \FOR{$q=1$ {\bfseries to} $1..h$}
                \FOR{$m=1$ {\bfseries to} $1..c$}
                    \FOR{$n=1$ {\bfseries to} $1..C$}
                        \STATE O(i, j, m) += I(i+x, j+y, n) * K(m, x, y, n)
                    \ENDFOR
                \ENDFOR
            \ENDFOR
         \ENDFOR
     \ENDFOR
   \ENDFOR
\end{algorithmic}
\end{algorithm}

Although the convolution operation involves several factors like kernel size, input size, and step, etc., the basic operation is general matrix multiplication. In other words, each step of convolution is a linear combination of \textbf{scalars}. Capsule convolution is similar to general convolution except that each basic operation linear combination of \textbf{tensors}. The problem is that we can not apply a linear combination to capsule convolution directly since input tensors and output tensors are of different dimensions. For example, we have an input $I(W, H, C, \mathbf{T_1})(D_1, D_2)$ with width $W$, height $H$, channel $C$ and tensor dimension $(D_1, D_2)$ at each location $(x, y)$. Also, we have a capsule kernel $K(w, h, c, \mathbf{T_2}(D_2, D_3))$ with width $w$, height $h$, channel $c$ and tensor dimension $(D_2, D_3)$. Finally, we are supposed to have an output $O(W', H', C', \mathbf{T_3})(D_1, D_3)$.
Note that $T_1$ and $T_2$ are tensors with rank $> 1$ as long as they can `inner product' with each other. For simple demonstration, we assume both $T_1$ and $T_2$ are metrics. A general capsule convolution can be calculated as in Algorithm~\ref{alg:cap_conv}.

\begin{algorithm*}[tb] 
   \caption{General capsule convolution.}
   \label{alg:cap_conv}
\begin{algorithmic}
   \STATE {\bfseries Input:} $I(W, H, C, \mathbf{T_1}(D1, D2))$
   \STATE {\bfseries Kernel:} $K(w, h, c, \mathbf{T_2}(D_2, D_3))$
       \STATE {\bfseries Output:} $O(W', H', C', \mathbf{T_3}(D_1, D_3))$
   \FOR{$p=1$ {\bfseries to} $1..C'$}
      \FOR{$i=1$ {\bfseries to} $1..H'$}
         \FOR{$j=1$ {\bfseries to} $1..W'$}
            \STATE $row\_offset = i * stride$
            \STATE $col\_offset = j * stride$
            \FOR{$s=1$ {\bfseries to} $1..C$}
                \FOR{$m=1$ {\bfseries to} $1..h$}
                    \FOR{$n=1$ {\bfseries to} $1..w$}                    
                        \FOR{$c=1$ {\bfseries to} $1..D_1$}
                            \STATE ${idx} = (p * C * h * w + s * h * w + m * w + n) * D_1 * D_2 * D_3 + c * D_2 * D_3$
                            \STATE $i_{idx} = (s * H * W + (row\_offset + m) * W + col\_offset + n) * D_1 * D_2 + c * D_1 * D_2$
                            \STATE $o_{idx} =(p * H' * W' + i * W' + j) * D_1 * D_3 +                          c * D_1 * D_3$
                            \STATE $\mathbf{matrix\_ multiply} (I+i_{idx}, K + idx, O + o_{idx})$
                            
                        \ENDFOR
                    \ENDFOR
                \ENDFOR
            \ENDFOR
         \ENDFOR
     \ENDFOR
   \ENDFOR
\end{algorithmic}
\end{algorithm*}

\subsection{Accelerating capsule convolutions}
By comparing Algorithm~\ref{alg:conv} and Algorithm~\ref{alg:cap_conv}, we can see that the fundamental difference between general convolution and capsule convolution is that general convolution involves only one linear combination at each step while capsule convolution involves kernel-size times of matrix multiplications and one time of linear combination among matrics. Of course, we can always implement capsule convolutions by calling the convolution interfaces multiple times (that's how most CapsNets are implemented), but this is the exact reason for the inefficiency. A reasonable idea would be paralleling matrix multiplications (during transforming input tensors to be the same shape as output tensors) as well as tensor combinations. To achieve that, we can simply lay out all the local processes first, then re-organize them in a way we can take advantage of the current CUDA APIs, and finally collect the result. To spread all the local processes, we adopt a similar idea of im2col function in Matlab, but with a capsule version.

\section{Our acceleration algorithm}\label{section:acce}
\begin {figure*}[t]
\begin{center} 
\includegraphics[width=.8\textwidth]{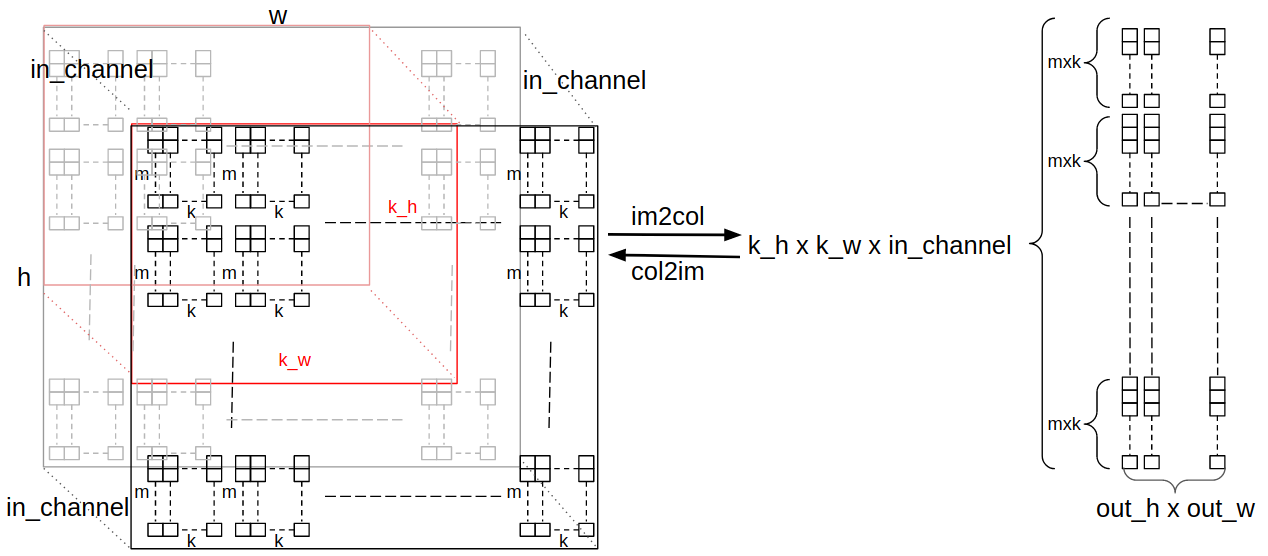}
\caption{Step\#1, we take k\_w * k\_h * (in\_channel * (m * k)) shape of data from the input (the shape is (w * h * in\_channel*(m*k)) to form one column. Eventually, we get output\_h * output\_w columns.}
\label{step_1}
\end{center}
\end{figure*}

\begin {figure*}[t]
\begin{center} 
\includegraphics[width=.8\textwidth]{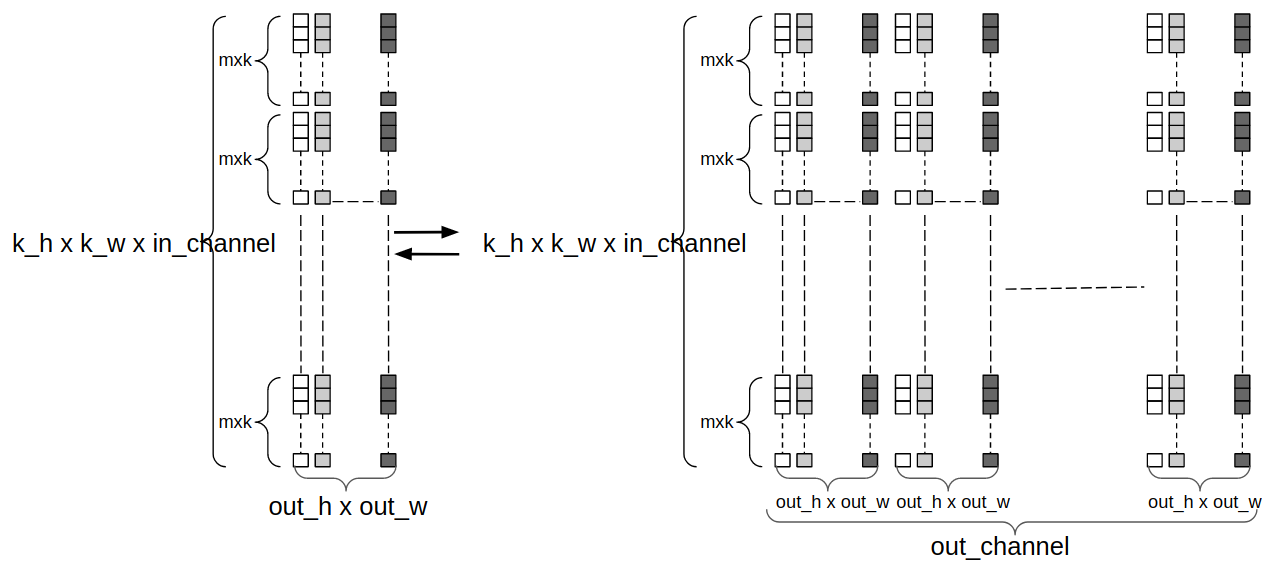}
\caption{Step\#2 replicates the input data, who has a shape of output\_h * output\_w * k\_w * k\_h * (in\_channel * (m * k)), for out\_channel times.}
\label{step_2} 
\end{center}
\end{figure*}

\begin {figure*}[t]
\begin{center} 
\includegraphics[width=.8\textwidth]{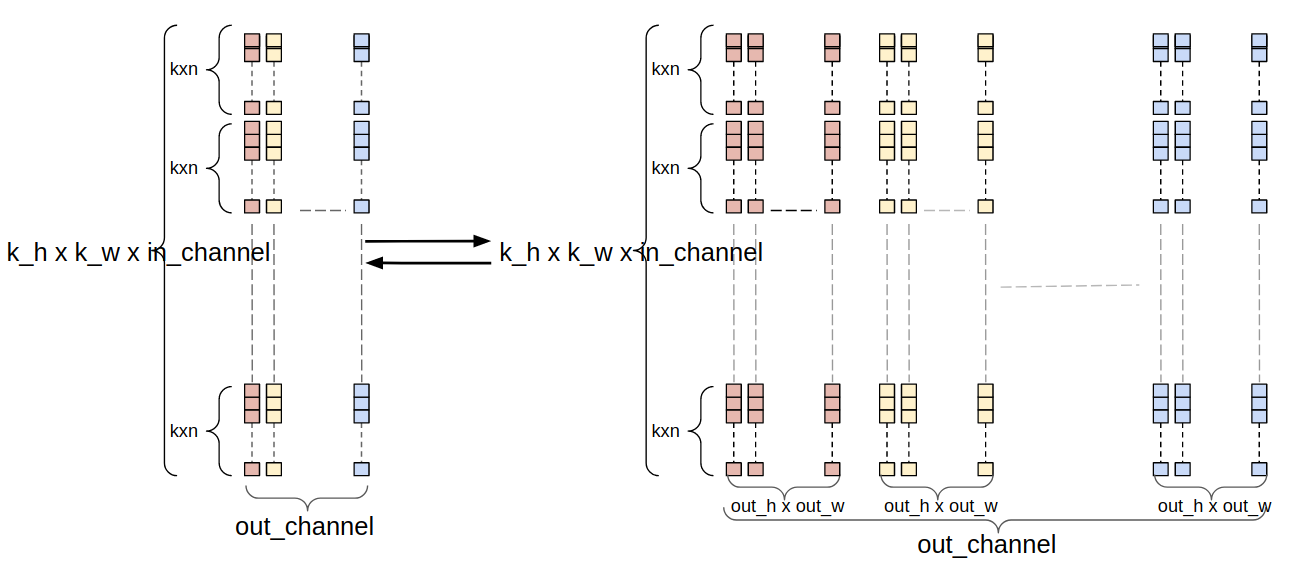}
\caption{Step \#3 replicates each column of the capsules (each column represents one channel in the output space.) for out\_h * out\_w times.}
\label{step_3} 

\end{center}
\end{figure*}

\begin {figure*}[t]
\begin{center} 
\includegraphics[width=.8\textwidth]{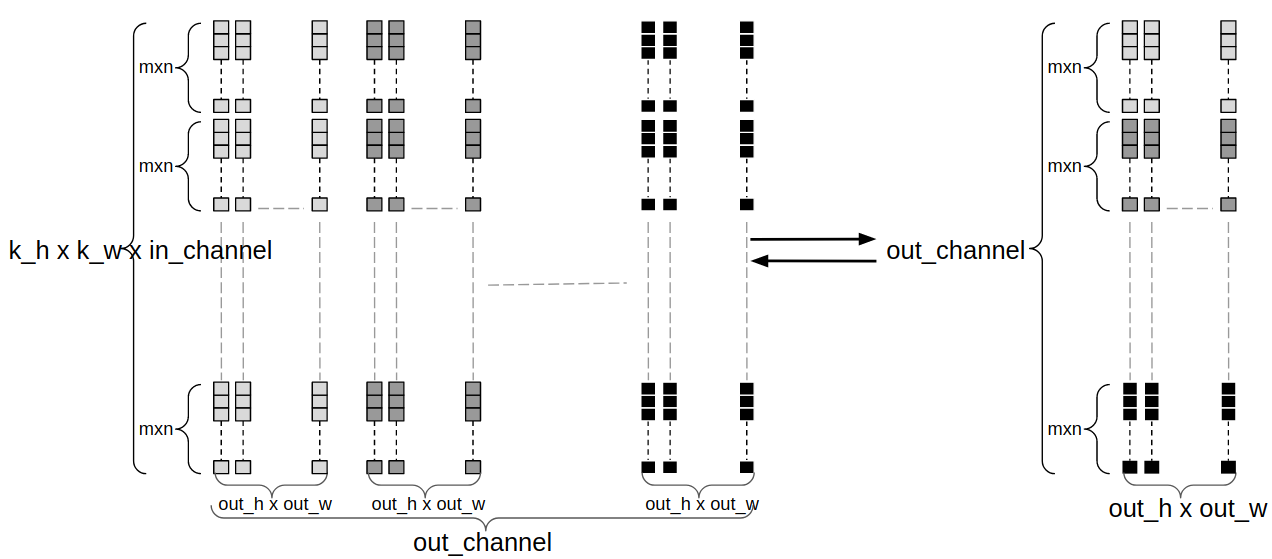}
\caption{Step \#4 applies batched multiplication to the extended input (from step\#2) and the extended capsules (from step\#3). Then a linear combination of tensors (the shape is (m, n)) are applied to acquire the final output.}
\label{step_4} 

\end{center}
\end{figure*}

The primary idea of our algorithm is pre-processing the input, output, and capsule kernels. Then we do the calculations in a parallel way and finally recover the output. We split our algorithm into forward pass (as shown in Algorithm~\ref{alg:acce_cap_conv_forward}) and backward pass (as shown in Algorithm~\ref{alg:acce_cap_conv_backward}). Below shows how each function in both passes work: 

\begin{itemize}
  \item $I_{flattened} = capsule\_im2col(I)$. This function pre-processes the input for step\#2 in parallel. Specifically, we collect kernel-size data for the input across all the strides and channels. Please check Figure~\ref{step_1} for examples.
  \item $I^{'} = input\_extend(I_{flattened})$. This function replicates the data that from the last step out\_channel times to prepare for step\#4. Please check Figure~\ref{step_2} for details.
  \item The $kernel\_extend(K)$ replicates each column of the capsules for out\_h * out\_w times to prepare for step\#4. Please check Figure~\ref{step_3} for details. 
  \item $strided\_batched\_matrix\_multiply(K', I')$ applies batched matrix multiplication. This function is available on CUDA~\cite{cuda}.
  \item $output\_reduce(O')$ recovers the output by applying batched, strided linear combinations among matrices, as Figure~\ref{step_4} shows. 
\end{itemize}

\begin{figure}[!]
  \centering
  \includegraphics[width=0.7\linewidth]{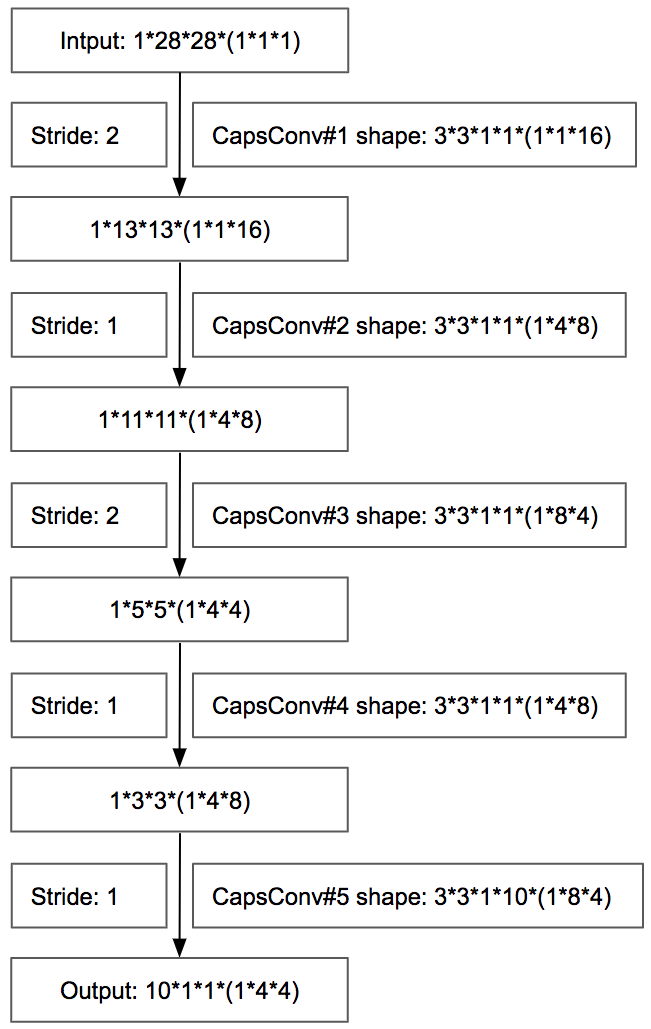}
  \caption{The sample P-CapsNets. }
  \label{pcapsnet_eg}
\end{figure}

\begin{table*}[t]
\caption{}
\label{result}
\vskip 0.15in
\begin{center}
\begin{small}
\begin{sc}
\begin{tabular}{lcccr}
\toprule
Acceleration Type & Time (ms) & forward (ms) & backward (ms) \\
\midrule
Ours with official APIs & 510 & 170  & 340\\
Ours with custom APIs  &  520 & 170  & 350\\
GPU &  2220 & 650  & 1570\\

\bottomrule
\end{tabular}
\end{sc}
\end{small}
\end{center}
\vskip -0.1in
\end{table*}

\begin{algorithm*}[tb] 
   \caption{Accelerating capsule convolution (forward).}
   \label{alg:acce_cap_conv_forward}
\begin{algorithmic}
    \STATE {\bfseries Input:} $I(W, H, C, \mathbf{T_1}(D_1, D_2))$
    \STATE {\bfseries Kernel:} $K(w, h, c, \mathbf{T_2}(D_2, D_3))$
    \STATE {\bfseries Output:} $O(W', H', C', \mathbf{T_3}(D_1, D_3))$
    \STATE $I_{flattened} = capsule\_im2col(I)$
    \STATE $I' = input\_extend(I_{flattened})$
    \STATE $K' = kernel\_extend(K)$
    \STATE $O' = strided\_batched\_matrix\_multiply(K', I')$
    \STATE $ O = output\_reduce(O')$
    
\end{algorithmic}
\end{algorithm*}

\begin{algorithm*}[tb] 
   \caption{Accelerating capsule convolution (backward).}
   \label{alg:acce_cap_conv_backward}
\begin{algorithmic}
    \STATE {\bfseries Input:} $I(W, H, C, \mathbf{T_1}(D_1, D_2))$
    \STATE {\bfseries Kernel:} $K(w, h, c, \mathbf{T_2}(D_2, D_3))$
    \STATE {\bfseries Output\_diff:} $O{d}(W', H', C', \mathbf{T_3}(D_1, D_3))$
    \STATE {\bfseries Input\_diff:} $I{d}(W, H, C, \mathbf{T_1}(D_1, D_3))$
    \STATE {\bfseries Kernel\_diff:} $K_{diff}(w, h, c, \mathbf{T_2}(D_2, D_3))$

    \STATE $I_{O_{d}^{'}} = output\_extend(O_{d})$
    \STATE $K_{diff}^{'} = strided\_batched\_matrix\_multiply(O_{d}^{'}, I')$
    \STATE $K_{diff} = strided\_batched\_matrix\_multiply(K_{diff}^{'})$
    \STATE $K' = kernel\_extend(K)$
    \STATE $I_{d}^{'} = strided\_batched\_matrix\_multiply(K', O')$
    \STATE $I_{d} = input\_reduce(K', O')$
    \STATE $I_{d} = capsule\_col2im(I_{d})$

\end{algorithmic}
\end{algorithm*}

\begin{algorithm*}[tb] 
   \caption{Accelerating capsule convolution (forward, no CUDA matrix multiplication interface called).}
   \label{alg:acce_cap_conv_forward_2}
\begin{algorithmic}
    \STATE {\bfseries Input:} $I(W, H, C, \mathbf{T_1}(D_1, D_2))$
    \STATE {\bfseries Kernel:} $K(w, h, c, \mathbf{T_2}(D_2, D_3))$
    \STATE {\bfseries Output:} $O(W', H', C', \mathbf{T_3}(D_1, D_3))$
    \FORALL{index}
    \STATE $i_{idx} = calculate\_input\_index(index)$
    \STATE $o_{idx} = calculate\_output\_index(index)$
    \STATE $idx = calculate\_weight\_index(index)$
    \STATE $\mathbf{matrix\_ multiply} (I+i_{idx}, K + idx, O + o_{idx})$
    \ENDFOR

\end{algorithmic}
\end{algorithm*}

As Algorithm~\ref{alg:acce_cap_conv_forward} and Algorithm~\ref{alg:acce_cap_conv_backward} show, the key point of our algorithm is paralleling as many matrix multiplication operations as possible and at the same time minimizing the time of preparing these matrices. The next question is that can we write our own paralleling algorithms instead of using the strided, batched matrix APIs in CUDA? 
Based on this idea, we pre-calculated three indices of all the matrix multiplication operations for the input, capsules, and output, then spread all the operations into a parallel for-loop. Finally, collect the result by accumulating all the products at each output position. Algorithm~\ref{alg:acce_cap_conv_forward_2} shows how the forward pass works. We found that there is no big difference between this scheme and the proposed one ( as Table~\ref{result} shows). Please check our code for details.   

\section{Experiments}
To focus on capsule convolutions, we build a 5-layer convolutional CapsNets, as Figure~\ref{pcapsnet_eg} shows rather than use the classic CapsNets~\cite{emrouting, dyrouting}. All our experiments are based on a TITAN X (Pascal) GPU with a CUDA version of 10.0. As Table~\ref{result} shows, our two solutions achieve 4x times acceleration.

\section{Conclusion}
We focus on how to accelerate capsule networks that has been neglected by most CapsNets researchers. Traditional CapsNets usually involves one layer of fully-connected capsule layer, which already make it slower than normal CNNs. Accelerating only Routing procedures is not enough, especially for a network that contains multiple capsule layers. That's probably the reason it is difficult to build a 50 or 100 capsule layers network. Our acceleration algorithm provides a potential solution for very deep CapsNets.  

\nocite{langley00}

\bibliography{main}

\begin{thebibliography}{6}
\providecommand{\natexlab}[1]{#1}
\providecommand{\url}[1]{\texttt{#1}}
\expandafter\ifx\csname urlstyle\endcsname\relax
  \providecommand{\doi}[1]{doi: #1}\else
  \providecommand{\doi}{doi: \begingroup \urlstyle{rm}\Url}\fi

\bibitem[Choi et~al.(2019)Choi, Seo, Im, and Kang]{attention_routing}
Choi, J., Seo, H., Im, S., and Kang, M.
\newblock Attention routing between capsules.
\newblock \emph{CoRR}, abs/1907.01750, 2019.
\newblock URL \url{http://arxiv.org/abs/1907.01750}.

\bibitem[Hinton et~al.(2018)Hinton, Sabour, and Frosst]{emrouting}
Hinton, G.~E., Sabour, S., and Frosst, N.
\newblock Matrix capsules with {EM} routing.
\newblock In \emph{International Conference on Learning Representations}, 2018.
\newblock URL \url{https://openreview.net/forum?id=HJWLfGWRb}.

\bibitem[Li et~al.(2018)Li, Guo, Dai, Ouyang, and Wang]{encapsule}
Li, H., Guo, X., Dai, B., Ouyang, W., and Wang, X.
\newblock Neural network encapsulation.
\newblock \emph{ECCV}, 2018.

\bibitem[NVIDIA et~al.(2020)NVIDIA, Vingelmann, and Fitzek]{cuda}
NVIDIA, Vingelmann, P., and Fitzek, F.~H.
\newblock Cuda, release: 10.2.89, 2020.
\newblock URL \url{https://developer.nvidia.com/cuda-toolkit}.

\bibitem[Sabour et~al.(2017)Sabour, Frosst, and Hinton]{dyrouting}
Sabour, S., Frosst, N., and Hinton, G.~E.
\newblock Dynamic routing between capsules.
\newblock \emph{CoRR}, abs/1710.09829, 2017.
\newblock URL \url{http://arxiv.org/abs/1710.09829}.

\bibitem[Zhang et~al.(2018)Zhang, Zhao, Wu, and Zhou]{fast_dyrouting}
Zhang, S., Zhao, W., Wu, X., and Zhou, Q.
\newblock Fast dynamic routing based on weighted kernel density estimation.
\newblock \emph{CoRR}, abs/1805.10807, 2018.
\newblock URL \url{http://arxiv.org/abs/1805.10807}.

\end{thebibliography}
\bibliographystyle{icml2021}

\end{document}